%% file: template.tex
\documentclass{article}

\usepackage{arxiv}

\usepackage[utf8]{inputenc} 
\usepackage[T1]{fontenc}    
\usepackage{hyperref}       
\usepackage{url}            
\usepackage{booktabs}       
\usepackage{amsfonts}       
\usepackage{nicefrac}       
\usepackage{microtype}      
\usepackage{lipsum}
\usepackage{graphicx}
\graphicspath{ {./images/} }
\usepackage{array,multirow}
\usepackage{authblk}
\usepackage{amsmath}
\usepackage{amssymb}
\usepackage{mathtools}
\usepackage{dirtytalk}

\newtheorem{theorem}{Theorem}[section]

\newtheorem{definition}[theorem]{Definition}

\title{Differentially Private Graph Classification with GNNs}

\author[1,2]{Tamara T. Mueller}
\author[3,6]{Johannes C. Paetzold}
\author[4]{Chinmay Prabhakar}
\author[1,2]{Dmitrii Usynin}
\author[1,2,5]{Daniel Rueckert}
\author[1,2,5]{Georgios Kaissis}

\affil[1]{Chair for AI in Medicine and Healthcare, Department of Informatics, Technical University of Munich}
\affil[2]{Department of Diagnostic and Interventional Radiology, Faculty of Medicine, Technical University of Munich}
\affil[3]{Department of Informatics, Technical University of Munich}
\affil[4]{Department of Quantitative Bio Medicine, University of Zurich}
\affil[5]{Department of Computing, Imperial College London}
\affil[6]{Institute for Tissue Engineering and Regenerative Medicine, Helmholtz Zentrum München}

\begin{document}
\maketitle
\begin{abstract}
Graph Neural Networks (GNNs) have established themselves as the state-of-the-art models for many machine learning applications such as the analysis of social networks, protein interactions and molecules. Several among these datasets contain privacy-sensitive data. Machine learning with differential privacy is a promising technique to allow deriving insight from sensitive data while offering formal guarantees of privacy protection. However, the differentially private training of GNNs has so far remained under-explored due to the challenges presented by the intrinsic structural connectivity of graphs. In this work, we introduce differential privacy for graph-level classification, one of the key applications of machine learning on graphs. Our method is applicable to deep learning on multi-graph datasets and relies on differentially private stochastic gradient descent (DP-SGD). We show results on a variety of synthetic and public datasets and evaluate the impact of different GNN architectures and training hyperparameters on model performance for differentially private graph classification. Finally, we apply explainability techniques to assess whether similar representations are learned in the private and non-private settings and establish robust baselines for future work in this area.
\end{abstract}


\input{01_introduction}
\input{02_relatedwork}

\input{03_theory}
\input{04_experiments}

\input{05_results}

\input{06_conclusion}

\bibliographystyle{unsrt}  
\bibliography{literature}  

\newpage
\appendix
\onecolumn
\section{Appendix}

\input{07_appendix}

\end{document}

%% file: 01_introduction.tex
\section{Introduction}
\label{sec:intro}
The introduction of geometric deep learning, and more specifically Graph Neural Networks (GNNs) \cite{gori2005new} has enabled training ML models on data in non-Euclidean spaces, with state-of-the-art performance in many applications. GNNs are able to directly leverage the graph structure of the data and propagate the information stored in nodes of the graph along the edges connecting nodes with each other. Thus, the information flow through the network respects the underlying topology of the graph. 

\begin{figure}
    \centering
    \includegraphics[width=0.8\textwidth]{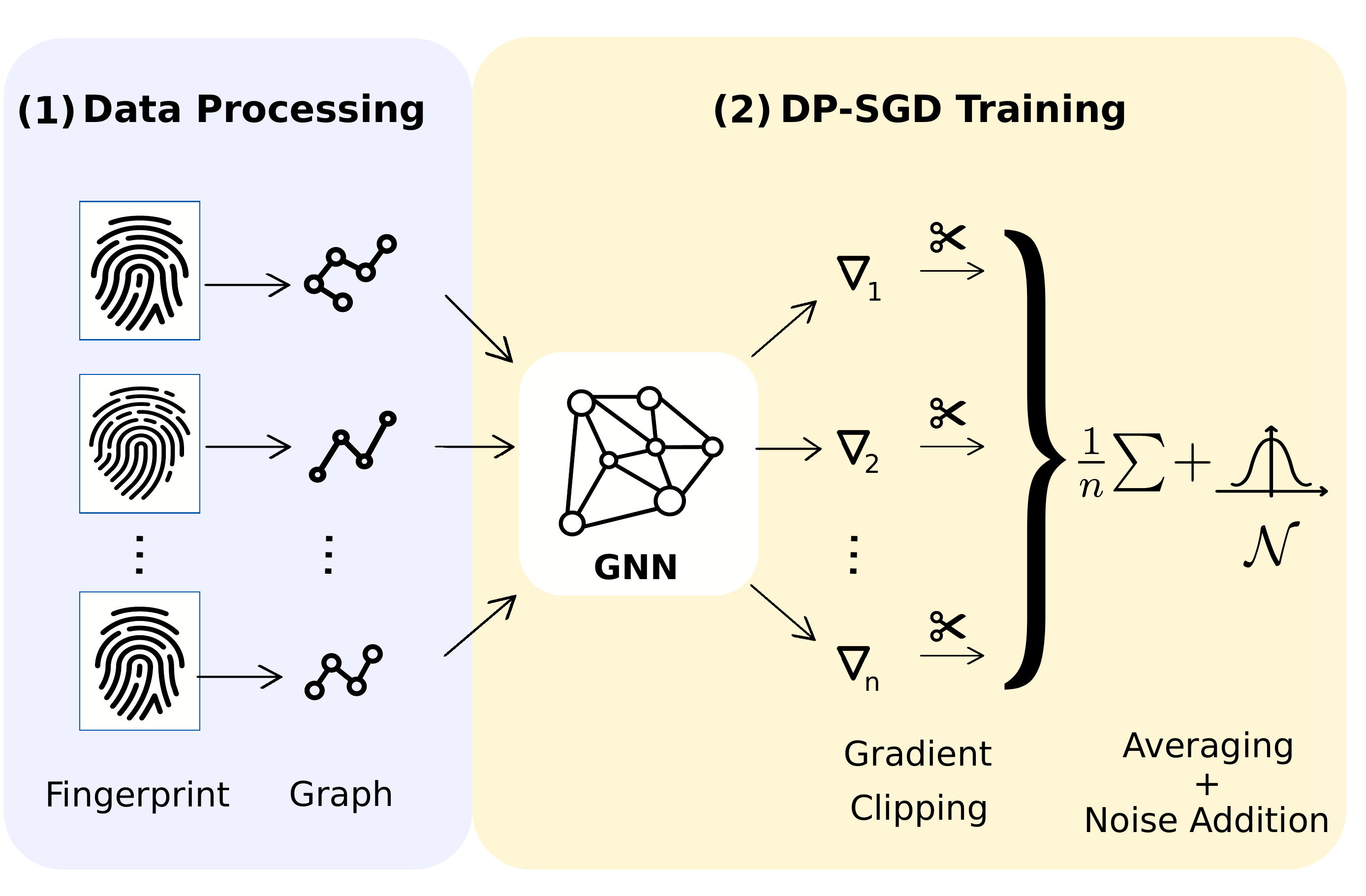}
    \caption{Overview of our differentially private training method for graph classification on a fingerprint dataset. In step (1) the fingerprint images are converted into graphs, which are then in step (2) passed to a GNN model, which is trained with differentially private stochastic gradient descent (DP-SGD). The individual gradients are clipped, then averaged and Gaussian noise is added.}
    \label{fig:overview}
\end{figure}

In general, GNNs have been employed in three types of problem areas: node classification, edge prediction, and graph classification. In this work, we focus on graph classification tasks. In the setting of graph classification (also termed graph property prediction), the dataset consists of multiple graphs and a GNN is trained to predict one label for each individual graph, predicting a specific property of the whole graph. Application areas of geometric deep learning range from social networks \cite{fan2019graph} to medical applications \cite{li2019graph,mao2019medgcn}, drug discovery or molecule classification \cite{duvenaud2015convolutional}, spatial biological networks \cite{paetzold2021whole} and shape analysis \cite{wei2020view}. Drawing meaningful insights in many of these application areas fundamentally relies upon the utilisation of privacy-sensitive, often scarce, training  data belonging to individuals. For example when using functional magnetic resonance imaging (fMRI) for identifying disease-specific biomarkers of brain connectivity like in \cite{li2019graph} and \cite{li2021braingnn}, the graph data encodes sensitive, patient-specific medical data.

The reliance on sensitive data in machine learning holds potential for misuse and can therefore be associated with the risks to individual participants' privacy. Various machine learning contexts have been shown vulnerable to be exploited by malicious actors, resulting in a leakage of private attributes \cite{ganjuproperty2018}, of membership information \cite{shokri2017membership} or even in full dataset reconstruction \cite{zhang2019secret,geiping2020inverting}. In graph machine learning, the data and the models trained on that data are \textit{by design} more vulnerable to adversarial attacks targeting privacy of the data owners. This is attributed to the fact that graphs incorporate additional information that is absent from typical Euclidean training contexts, such as the relational information about the nodes in the graph. This auxiliary, highly descriptive information can be leveraged by an adversary to assist them in sensitive information extraction, which has been demonstrated in a number of prior works \cite{zhang2021graphmi,he2021node,olatunji2021membership}. Such attacks can also be facilitated by the choice of learning context in cases the model is trained collaboratively. For instance, transductive collaborative learning renders attacks aimed at disclosing the membership of individual training points trivial \cite{he2021node}. Of note, such additional information embedded in graphs is often essential for effective GNN training and is, thus, non-trivial to privatise or remove, as it would be highly detrimental to the performance of the model. 

It is thus apparent that the implementation of privacy-enhancing techniques is required to facilitate the training of models of sensitive graph-structured data, but such techniques must also respect the particularities of graph machine learning. Our work utilises a formal method of privacy preservation termed differential privacy (DP) \cite{dwork2014algorithmic} which, when applied to machine learning training, is able to objectively quantify the privacy loss for individual input data points. DP methods have been successfully applied to numerous problems such as medical image analysis \cite{ziller2021medical, kaissis2021end}, natural language processing (NLP) \cite{basu2021benchmarking}, 
reinforcement learning \cite{zhuo2019federated} or generative models \cite{frigerio2019differentially} and have shown promising results. DP guarantees that the information gain from observing the output of an algorithm trained on datasets differing in one individual is (sometimes with high probability), bounded by a (typically small) constant.

In this work, motivated by the above-mentioned requirements for objective privacy guarantees in machine learning tasks involving graph-structured data, we study the problem of efficient differentially private graph neural network training for graph classification tasks. We utilise differentially private stochastic gradient descent (DP-SGD) \cite{abadi2016deep}, a technique tailored to neural network training, which, due to its compatibility with existent deep learning workflows, can be seamlessly integrated and therefore offers high generalisability to new model architectures and problem spaces. We show that DP-SGD can be applied to graph learning and evaluate our results with respect to privacy budgets and network performance on four different datasets. To the best of our knowledge, this is the first work that shows the application of differentially private GNNs for graph classification tasks. Combined with our investigation of the explainability technique \textit{GNNExplainer} to determine differences between DP and non-DP models, this work can serve as a baseline for future work in this area. Our contributions can be summarised as follows: 

\begin{enumerate}
    \item We formally extend the application of DP-SGD to graph classification tasks;
    \item We evaluate the differentially private training of commonly utilised graph neural networks on a number of benchmark and real-world datasets and investigate the effects of DP training on model utility and privacy guarantees;
    \item To assess whether privately and non-privately trained models learn similar representations, we apply \mbox{GNNExplainer}, a state-of-the-art explainability technique tailored to graph neural networks.
\end{enumerate}

%% file: 02_relatedwork.tex
\section{Related Work}
\label{sec:related_work}
Specific facets of differentially private graph analysis have been addressed in prior work: Since the introduction of differentially private computation on graph data in 2007 by Nissim et al. \cite{nissim2007smooth}, \textit{node-level} and \textit{edge-level} DP have been established as the two DP formalisms on graphs \cite{DPGraphsSlides}. As discussed in the Theory section, the definition of DP relies on the notion of adjacent datasets, that is, datasets differing in the data of one individual. In the setting of tabular data for example, two datasets are adjacent if they differ in one row. In node-level DP, two graph datasets are interpreted as adjacent if one node and its incident edges is inserted or removed. For edge-level DP, on the other hand, two datasets are regarded as adjacent if they differ in exactly one edge. Therefore, node-level DP is a strictly stronger privacy guarantee in comparison to edge-level DP \cite{kasiviswanathan2013analyzing}. As real-world graphs are prevalently sparse, removal of a single node can severely alter the graph's structure \cite{kasiviswanathan2013analyzing}, whereas removal of an edge typically has a less severe impact on the resulting graph structure. 

Implementations of the aforementioned techniques have been presented in the context of graph neural network training. For instance, Igamberdiev et al. \cite{igamberdiev2021privacy} explore the application of DP on Graph Convolutional Networks (GCNs) \cite{kipf2016semi} for node classification. They evaluate privacy guarantees for text classification on benchmark datasets and achieve rigorous privacy guarantees while maintaining high model performance. Daigavan et al. \cite{daigavane2021node} formalise the notion of node-level DP on one-layer GNNs with an extension of privacy amplification by sampling to GNNs and evaluate their method on several benchmark datasets in node classification tasks. Different approaches to the here introduced application of differential privacy have been explored in the context of federated learning on graphs and locally private graph neural network training. Zhou et al. \cite{zhou2020vertically}, for example, introduce a vertically federated GNN for node classification tasks and Sajadmanesh et al. \cite{sajadmanesh2021locally} introduce a framework to train locally private GNNs.

However, to our knowledge, the application of DP algorithms specifically to graph property prediction has neither been formalised nor evaluated.

%% file: 03_theory.tex
\section{Theory}
In this section, we introduce and formalise the theory to train graph neural networks for graph property prediction using the concept of differentially private stochastic gradient descent (DP-SGD). 

\subsection{GNNs for Graph Property Prediction}
The objective of graph classification (also known as graph property prediction) is to predict a specific property of interest for an entire graph $\mathcal{G}$. $\mathcal{G}$ is an unweighted and undirected graph with $\mathcal{G}=(\mathcal{V}, \mathcal{E})$, where $\mathcal{V}$ is the set of nodes and $\mathcal{E}$ is the set of all edges of the graph. For the experiments in this work, we will use three commonly used GNN models: Graph Convolutional Networks (GCNs) \cite{kipf2016semi}, Graph Attention Networks (GATs) \cite{velivckovic2017graph}, and GraphSAGE \cite{hamilton2017inductive}.

\subsection{Differential Privacy}
Differential Privacy (DP) \cite{dwork2014algorithmic} is a theoretical framework and collection of techniques aimed at enabling analysts to draw conclusions from datasets while safeguarding individual privacy. Intuitively, an algorithm preserves DP if its outputs are approximately equivariant to the inclusion or exclusion of a single individual in the dataset over which the algorithm is executed. The DP guarantee is given in terms of probability mass/density of the algorithm's outputs.

In the current study, we assume that an analyst $\mathcal{A}$ is entrusted with a multi-graph database $D$ of cardinality $N$ containing privacy-sensitive graphs $\mathcal{G}_i \in D, i \in [1, \cdots, N]$ by a group of individuals. We assume that each individual's graph is only present in the database once. From $D$, an \textit{adjacent} database $D'$ of cardinality $N\pm1$ can be constructed by adding or removing a single individual's graph. We denote adjacency by $D\simeq D'$. The set (\textit{universe}) of all adjacent databases forms a metric space $X$ with associated metric $d_X$, in our case, the Hamming metric. 

We additionally assume that $\mathcal{A}$ executes a \textit{query function} $f$ over an element of $X$. In our study, the application of $f$ represents a sequential composition of the forward pass, loss calculation and gradient computation of a graph neural network for each individual input (training example) to $f$. We then define the $L_2$-sensitivity of $f$ as follows:

\begin{definition}[$L_2$-sensitivity of $f$]
Let $f, X$ and $d_X$ be defined as above. Additionally, let $Y$ be the metric space of $f$'s outputs with associated metric $d_Y$. When $Y$ is the Euclidean space and $d_Y$ the $L_2$ metric, we define the $L_2$-sensitivity $\Delta$ of $f$ as:
\begin{equation}
    \Delta := \max_{D,D' \in X, D \simeq D'} \frac{d_Y(f(D), f(D'))}{d_X(D,D')}.
\end{equation}
We remark that the maximum is taken over all adjacent database pairs in $X$. Moreover, $\Delta$ describes a Lipschitz condition on $f$, implying that $\Delta \equiv K_f$, where $K_f$ is the Lipschitz constant of $f$. This in turn implies that $\Delta = \sup \Vert \nabla f \Vert_2$. In our case, the $L_2$-sensitivity of the loss function therefore corresponds to the upper bound on its gradient.
\end{definition}
\noindent
We can now define the Gaussian Mechanism on $f$:

\begin{definition}[Gaussian Mechanism]
Let $f, \Delta$ be defined as above. The Gaussian mechanism $\mathcal{M}$ operates on the outputs of $f$, $\mathbf{y}=f(x)$, where $\mathbf{y} \in \mathbb{R}^n$ as follows:
\begin{equation}
    \mathcal{M}(\mathbf{y}) = \mathbf{y} + \xi,
\end{equation}
where $\xi \sim \mathcal{N}(0, \sigma \mathbb{I}^n)$, $\sigma$ is calibrated to $\Delta$, and $\mathbb{I}^n$ is the identity matrix with $n$ diagonal elements.
\end{definition}
\noindent
When $\sigma$ is appropriately calibrated to $\Delta$, $\mathcal{M}$ preserves $(\varepsilon, \delta)$-DP:

\begin{definition}[$(\varepsilon, \delta)$-DP]
$\mathcal{M}$ preserves $(\varepsilon, \delta)$-DP if, $\forall \, S \subseteq \operatorname{Range}(\mathcal{M})$ and all adjacent databases $D, D'$ in $X$:
\begin{equation}
    \mathbb{P}(\mathcal{M}(D) \in S) \leq e^{\varepsilon} \mathbb{P}(\mathcal{M}(D') \in S) + \delta.
\end{equation}
We remark that the definition is symmetric.
\end{definition}

\subsection{DP-SGD}
Abadi et al. \cite{abadi2016deep} introduced an extension to stochastic gradient descent (SGD), termed DP-SGD to enable the differentially private training of neural networks. Here, at each training step, the Gaussian Mechanism is used to privatise the individual gradients of each training example before the model parameters are updated. However, since the sensitivity of the loss function in deep neural networks is -- in general -- unbounded, the gradient $L_2$-norm of each individual training example is \textit{clipped}, that is, projected to an $L_2$-ball of a pre-defined radius to artificially induce a bounded sensitivity condition before noise is applied. Tracking the privacy expenditure over the course of training (\textit{privacy accounting}) is enabled through the \textit{composition} property of DP, stating that repeated application of DP algorithms over the same data predictably degrades the privacy guarantees. In our study, a relaxation of DP termed Rényi DP (RDP) \cite{mironov2017renyi} is used for privacy accounting, due to its favourable compositional properties. RDP guarantees can be converted to $(\varepsilon, \delta)$-DP.

%% file: 04_experiments.tex
\section{Experiments}
\label{sec:experiments}

\begin{table}
\footnotesize
\begin{center}
\begin{tabular}{lcccc} 
\textbf{Dataset} & \shortstack{\textbf{(mean)}\\\textbf{\# nodes}} & \textbf{\# graphs} & \shortstack{ \textbf{\# node}\\\textbf{features}} & \# \textbf{classes} \\ [0.5ex] 
 \midrule
 \textbf{Synthetic} & 20 & 1000 & 9 & 2 \\
 \midrule
 \textbf{Fingerprints} & 7.6 & 1900 & 2 & 4 \\
  \midrule
 \textbf{Molbace} & 34 & 1513 & 9 & 2 \\
 \midrule
 \textbf{ECG} & 12 & 1125 & 512 & 2 \\ [1ex] 
\end{tabular}
 \caption{Overview of the utilised datasets and their characteristics. We report the mean number of nodes, in case the dataset contains graphs of varying sizes.} 
\label{tab:datasetoverview}
\end{center}
\end{table}

\subsection{Datasets}
We evaluate the application of DP-SGD in the context of graph property prediction tasks on four datasets. We rely on three publicly available and one synthetic dataset, generated to provide a reproducible and easy to control proof-of-concept. The three benchmark datasets tackle the problems of molecule classification (Molbace), fingerprint classification, and Left Bundle Branch Block (LBBB) detection on electrocardiogram (ECG) data. Table \ref{tab:datasetoverview} provides an overview of the datasets and their characteristics.

 \begin{figure}
    \centering
    \includegraphics[width=0.5\textwidth]{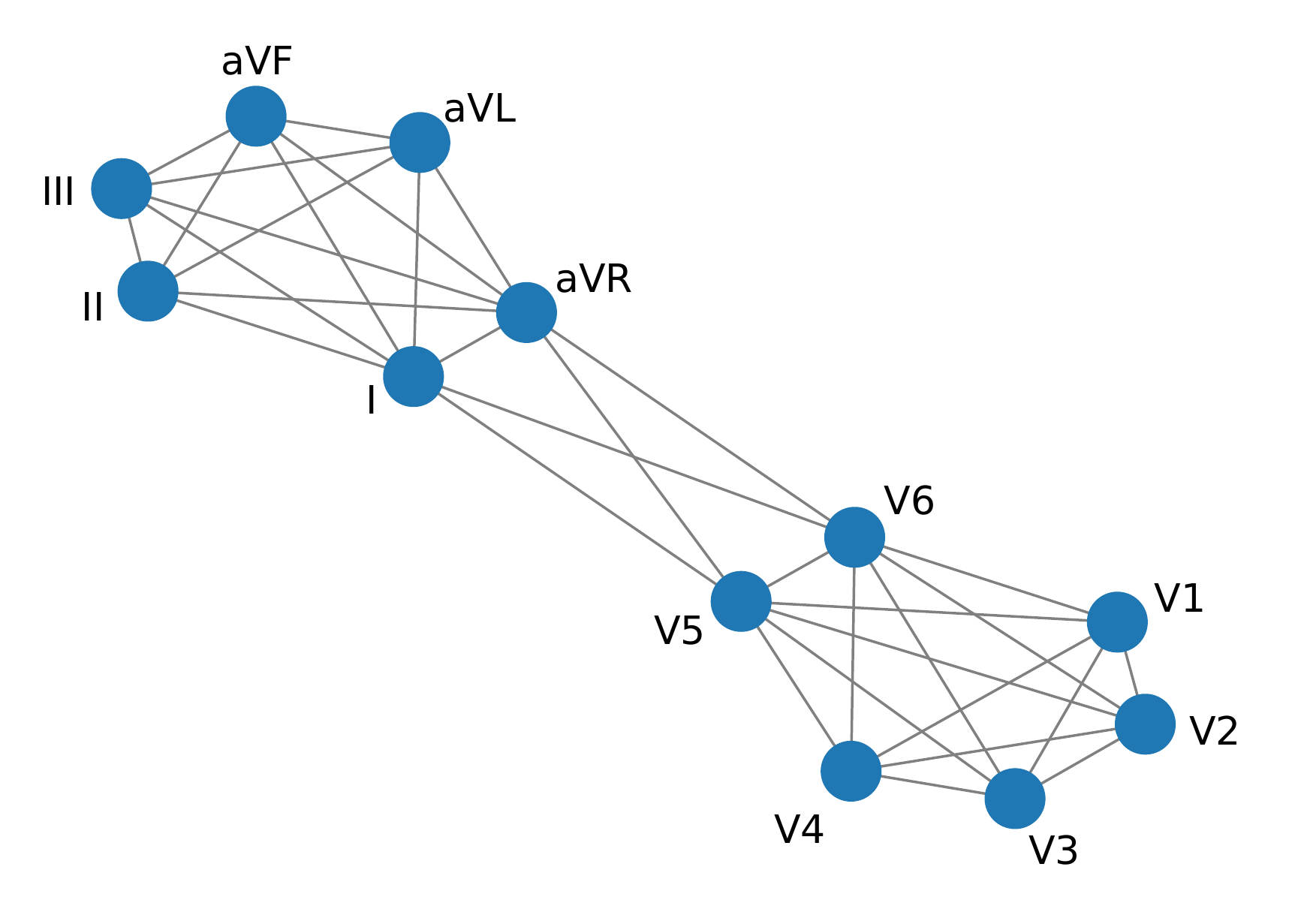}
    \caption{Graph visualisation of ECG data. We connected the different signal channels based on the medical location of the leads as well as prior knowledge. Leads I, II, III, aVF, aVL, and aVR are located on the extremities and the remaining leads on the chest.}
    \label{fig:ekg_data}
\end{figure}
 
\paragraph{Synthetic Dataset}
In order to derive a proof-of-concept of the novel application of DP-SGD on graph classification tasks, we construct a synthetic dataset, in which parameters can be manually controlled. We generate $1000$ individual Erd\H os-Rényi graphs of two classes. Each graph consists of twenty nodes and each node contains nine features, which are sampled from a normal distribution with the mean values of $0$ and $0.1$, while having the same standard deviation of $0.5$. The edge connection probabilities of the graphs from the two classes are set to $0.2$ and $0.3$, respectively. We generate $500$ graphs for each class to have a balanced dataset, split the dataset into $600$ training samples, $100$ validation samples and $300$ test samples, and perform binary graph classification. To address the applicability of our approach on graphs containing unconnected components, we do not eliminate unconnected nodes from the generated Erd\H os-Rényi graphs.


\paragraph{Fingerprints Dataset}
Fingerprint classification aims to separate images of fingerprints into different classes. A large within-class variability and a small separation between classes makes fingerprint classification a challenging task \cite{neuhaus2005graph}. We rely on the dataset introduced by Riesen et al. \cite{riesen2008iam} and provided by TU Datasets \cite{Morris_2020} to perform differentially private graph classification on fingerprints. The graphs are extracted from the images based on directional variance and the task follows the Galton-Henry classification scheme of five classes. We merge the five classes into four classes following the approach described in \cite{riesen2008iam}. Differentially private ML naturally befits this task, as it allows one to privatise the utilisation of the uniquely identifying fingerprint data for e.g. training machine learning models in tasks such as automated authentication.

\paragraph{Molbace Dataset}
To perform molecule classification in a binary graph classification setting, we use the benchmark dataset \textit{Molbace} from the OGB database \cite{hu2021open}, where the \textit{Molbace} dataset is adapted from MoleculeNet \cite{wu2018moleculenet}. It consists of $1513$ graphs, where each graph represents a molecule. Edges represent molecular bonds and nodes correspond to the atoms in the molecule. Each node contains $9$ node features and the average number of nodes per graph is 34. We split the dataset into $1210$ training graphs, $152$ test graphs and $151$ validation graphs. Node features contain atom features; for example the atomic number, chirality, formal charge, or whether the atom is in a ring or not. The prediction task of this dataset is to correctly classify whether the molecule inhibits HIV virus replication \cite{hu2021open}. Such a task is representative of federated learning workflows in which e.g. several pharmaceutical companies wish to jointly train a model for molecule property prediction, while wishing to keep their (possibly proprietary) molecule structures private from third parties.

\paragraph{ECG Dataset}
For the task of electrocardiogram (ECG) classification, we use the publicly available ECG dataset from the China Physiological Signal Challenge (CPSC) 2018 challenge dataset \cite{liu2018open}. We formulate a classification task between ECGs showing signs of a Left Bundle Branch Block and normal ECGs showing a sinus rhythm. LBBB is an insidious type of arrhythmia (that is, anomaly in the conduction system of the heart), which, when appearing suddenly, can herald acute myocardial ischemia or infarction. The ECG data consists of twelve ECG signal channels (\textit{leads}), recorded at different locations on the human torso and extremities. Leads affixed to the extremities constitute signal channels I, II, II, aVR, aVF and aVL. Leads affixed to the chest are used to derive signal channels V1 through V6. To construct a graph dataset from the ECG data, we utilise this medical motivation and divide the ECG extremity signal channels from the chest signal channels by fully connecting the extremity and chest subgraphs. In addition, we utilise prior knowledge about the leads which are typically used by physicians to delineate LBBB from sinus rhythm and thus connected channels I, aVR, V5, and V6. The structure of those graphs is visualised in Figure \ref{fig:ekg_data}. The dataset we use contains ECG data of $1125$ subjects. As ECG signals are periodic, we sub-sample the signals by only retaining the first $512$ signal points of each channel, leading to $512$ node features in the graphs. The binary classification dataset is highly imbalanced with $207$ subjects showing signs of LBBB and $918$ having normal ECG curves. Evidently, ECG data, like all medical data is highly sensitive, and thus requires formal methods of privacy protection.


\subsection{GNN Models for Graph Classification and DP-SGD Training}

Since the adoption of deep learning techniques to graph learning, most state-of-the-art methods for graph classification rely on a variant of \textit{message passing} to aggregate information across the nodes \cite{klicpera2018predict,chiang2019cluster,kong2020flag,huang2020combining,klicpera2019directional}. 

For our experiments, we implement a variety of GNN models to compare performance and evaluate the impact of DP on different graph learning techniques. We use GraphSAGE \cite{hamilton2017inductive}, Graph Attention Networks (GATs) \cite{velivckovic2017graph}, and Graph Convolutional Networks (GCNs) \cite{kipf2016semi}. For each dataset, we perform hyperparameter searches, leading to different models for each application. The depth of the GNNs varies from two to three layers with/without Instance Normalisation layers and with/without dropout, depending on the problem space. We do not use Batch Normalisation because of its incompatibility with differentially private training; Batch Normalisation, by taking averages across the batch during the forward pass, leaks information over samples in a batch and precludes the computation of \textit{per-sample} gradients necessary for DP-SGD. More details about the model architectures can be found in the supplementary material.

When training graph classification models with DP-SGD, we follow the standard procedure of DP-SGD training. Firstly, a privacy budget is set in terms of $\varepsilon$, then the model is trained with a specific noise multiplier that defines the amount of Gaussian noise added to the gradients of the model and a $L_2$-sensitivity bound. The model can then be trained a certain number of iterations, until the privacy budget $\varepsilon$ is reached. We then report the scores of the best-performing model out of the ones trained before the privacy budget is exhausted. For all differentially private training runs, we set $\delta=\frac{1}{N}$, where $N$ is the cardinality of the dataset and monitor the performance of the algorithm with different privacy budgets $\varepsilon$. Across all experiments, we utilise the same model architectures for DP-SGD and SGD training with the removal of potential dropout layers for DP-SGD training. In Table \ref{tab:results} we report the mean performance as well as the standard deviation of five independent runs for each experiment. We evaluate different scores for each model: ROC AUC, Accuracy, Sensitivity, Specificity and F1 Score. Hereby sensitivity reports the rate between the true positives and the sum of the true positives and false negatives. Specificity is the rate between the true negatives and the sum of the true negatives and false positives. The ROC AUC score is the Compute Area Under the Receiver Operating Characteristic Curve with a \textit{micro} average for multi-class datasets. Accuracy is the rate between the true positives and all samples and the F1 Score reports the harmonic mean of the precision and recall, also using a \textit{micro} averaging strategy for multi-class datasets.

%% file: 05_results.tex
\section{Experimental Results}
\label{sec:results}

\begin{table*}[!ht]
\centering
\scriptsize
\begin{tabular}{lll|ccccc|ccc}
\hline\hline
Data & Network & Training & ROC AUC & Accuracy & Sensitivity & Specificity & F1 Score & Noise & $L_2$-Clip & $\varepsilon$ \\
\hline\hline
\parbox[t]{5mm}{\multirow{4}{*}{\rotatebox[origin=c]{90}{\textit{Synthetic}}}} & GCN &  SGD & 0.934 $\pm$ 0.01 & 0.934 $\pm$ 0.01 & 0.955 $\pm$ 0.03 & 0.913 $\pm$ 0.03 & 0.934 $\pm$ 0.01  & - & - & - \\  
&  & DP-SGD & 0.918 $\pm$ 0.02 & 0.918 $\pm$ 0.02 & 0.897 $\pm$ 0.03 & 0.940 $\pm$ 0.02 & 0.917 $\pm$ 0.02 & 1.0 & 3.0 & 5 \\ 
&  & DP-SGD & 0.907 $\pm$ 0.02 & 0.910 $\pm$ 0.20 & 0.869 $\pm$ 0.04 & 0.946 $\pm$ 0.20 & 0.907 $\pm$ 0.02 & 2.0 & 3.0 & 1 \\  
&  & DP-SGD & 0.757 $\pm$ 0.11 & 0.756 $\pm$ 0.10 & 0.936 $\pm$ 0.06 & 0.575 $\pm$ 0.28 & 0.756 $\pm$ 0.10 & 2.2 & 3.0 &  0.5 \\  
\cline{2-11} 
& GAT & SGD & 0.912 $\pm$ 0.01 & 0.912 $\pm$ 0.01 & 0.940 $\pm$ 0.03 & 0.883 $\pm$ 0.06 & 0.912 $\pm$ 0.01 & -&-&- \\  
&  & DP-SGD & 0.893 $\pm$ 0.01 & 0.893 $\pm$ 0.01 & 0.895 $\pm$ 0.03 & 0.891 $\pm$ 0.03 & 0.893 $\pm$ 0.01 & 1.0 & 3.0 & 5 \\ 
&  & DP-SGD & 0.872 $\pm$ 0.02 & 0.872 $\pm$ 0.02 & 0.827 $\pm$ 0.04 & 0.907 $\pm$ 0.07 & 0.872 $\pm$ 0.02 & 2.0 & 3.0 & 1 \\ 
&  & DP-SGD & 0.575 $\pm$ 0.07 & 0.575 $\pm$ 0.07 & 0.730 $\pm$ 0.35 & 0.419 $\pm$ 0.42 & 0.576 $\pm$ 0.08 & 2.2 & 3.0 & 0.5 \\ 
\cline{2-11}
& SAGE &  SGD & 0.903 $\pm$ 0.02 & 0.902 $\pm$ 0.02 & 0.913 $\pm$ 0.04 & 0.893 $\pm$ 0.08 & 0.903 $\pm$ 0.02 & - & - & -   \\ 
&  & DP-SGD & 0.918 $\pm$ 0.01 & 0.918 $\pm$ 0.01 & 0.907 $\pm$ 0.03 & 0.933 $\pm$ 0.02 & 0.918 $\pm$ 0.01 & 1.0 & 3.0 &  5 \\ 
& &  DP-SGD & 0.893 $\pm$ 0.01 & 0.892 $\pm$ 0.01 & 0.872 $\pm$ 0.04 & 0.914 $\pm$ 0.03 & 0.893 $\pm$ 0.01 & 2.0 & 3.0 & 1 \\ 
& & DP-SGD & 0.598 $\pm$ 0.10 & 0.598 $\pm$ 0.11 & 0.609 $\pm$ 0.47 & 0.587 $\pm$ 0.39 & 0.598 $\pm$ 0.10 & 2.2 & 3.0 & 0.5 \\  
\hline\hline  
\parbox[t]{5mm}{\multirow{5}{*}{\rotatebox[origin=c]{90}{\textit{Fingerprint}}}} & GCN & SGD & 0.856 $\pm$ 0.01 & 0.785 $\pm$ 0.01 & 0.785 $\pm$ 0.01 & 0.928 $\pm$ 0.00 & 0.785 $\pm$ 0.01  &-&-& - \\ 
& & DP-SGD & 0.863 $\pm$ 0.01 & 0.794 $\pm$ 0.01 & 0.771 $\pm$ 0.01 & 0.932 $\pm$ 0.00 & 0.794 $\pm$ 0.01 & 1.0 & 3.0 & 5 \\ 
& & DP-SGD & 0.844 $\pm$ 0.02 & 0.766 $\pm$ 0.04 & 0.733 $\pm$ 0.05 & 0.921 $\pm$ 0.01 & 0.766 $\pm$ 0.04 & 1.8 & 3.0 & 1 \\ 
& & DP-SGD & 0.796 $\pm$ 0.06 & 0.693 $\pm$ 0.10 & 0.658 $\pm$ 0.09 & 0.898 $\pm$ 0.03 & 0.693 $\pm$ 0.09 & 2.3 & 3.0 & 0.5 \\   
\cline{2-11}
  & GAT & SGD & 0.857 $\pm$ 0.01 & 0.786 $\pm$ 0.01 & 0.764 $\pm$ 0.02 & 0.929 $\pm$ 0.01 & 0.786 $\pm$ 0.01 & - & - & -   \\ 
  &  & DP-SGD & 0.849 $\pm$ 0.02 & 0.774 $\pm$ 0.03 & 0.733 $\pm$ 0.04 & 0.924 $\pm$ 0.01 & 0.770 $\pm$ 0.03 & 1.0 & 3.0 & 5 \\  
  &  &  DP-SGD & 0.812 $\pm$ 0.02 & 0.728 $\pm$ 0.03 & 0.661 $\pm$ 0.01 & 0.906 $\pm$ 0.01 & 0.730 $\pm$ 0.03 & 1.8 & 3.0 & 1 \\ 
  &  &  DP-SGD & 0.737 $\pm$ 0.05 & 0.605 $\pm$ 0.08 & 0.585 $\pm$ 0.08 & 0.871 $\pm$ 0.03 & 0.610 $\pm$ 0.08 & 2.3 & 3.0 & 0.5 \\  
\cline{2-11}
& SAGE & SGD & 0.876 $\pm$ 0.02  & 0.814 $\pm$ 0.02 & 0.802 $\pm$ 0.03 & 0.940 $\pm$ 0.01 & 0.814 $\pm$ 0.02 & - & - & -   \\ 
& & DP-SGD & 0.869 $\pm$ 0.01 & 0.804 $\pm$ 0.01 & 0.788 $\pm$ 0.02 & 0.935 $\pm$ 0.01 & 0.804 $\pm$ 0.01 & 1.0 & 3.0 & 5 \\ 
& & DP-SGD & 0.861 $\pm$ 0.01 & 0.792 $\pm$ 0.01 & 0.776 $\pm$ 0.01 & 0.932 $\pm$ 0.00 & 0.791 $\pm$ 0.01 & 1.8 & 3.0 & 1 \\ 
& & DP-SGD & 0.712 $\pm$ 0.06 & 0.568 $\pm$ 0.08 & 0.529 $\pm$ 0.09 & 0.853 $\pm$ 0.03 & 0.568 $\pm$ 0.08 & 2.3 & 3.0 & 0.5 \\ [0.5ex]
\hline\hline  
\parbox[t]{5mm}{\multirow{3}{*}{\rotatebox[origin=c]{90}{\textit{ECG}}}} & GCN & SGD & 0.979 $\pm$ 0.01 & 0.932 $\pm$ 0.01 & 0.744 $\pm$ 0.03 & 0.979 $\pm$ 0.01 & 0.845 $\pm$ 0.02 & - & - & - \\  
 & & DP-SGD & 0.983 $\pm$ 0.01 & 0.904 $\pm$ 0.02 & 0.581 $\pm$ 0.07 & 0.983 $\pm$ 0.01 & 0.727 $\pm$ 0.06 & 0.6 & 5.0 & 10 \\ 
 & & DP-SGD & 0.983 $\pm$ 0.01 & 0.923 $\pm$ 0.01 & 0.644 $\pm$ 0.12 & 0.983 $\pm$ 0.01 & 0.772 $\pm$ 0.09  & 0.8 & 5.0 & 5\\ 
 & & DP-SGD & 0.986 $\pm$ 0.02 & 0.824 $\pm$ 0.03 & 0.169 $\pm$ 0.23 & 0.986 $\pm$ 0.02 & 0.231 $\pm$ 0.28 & 1.5 & 5.0 & 1 \\[0.5ex] 
\cline{2-11}
& GAT & SGD & 0.983 $\pm$ 0.01 &  0.922 $\pm$ 0.04 & 0.675 $\pm$ 0.19 & 0.983 $\pm$ 0.01 & 0.781 $\pm$ 0.17 & - & - & -    \\ 
& & DP-SGD & 0.968 $\pm$ 0.03 & 0.899 $\pm$ 0.01 & 0.637 $\pm$ 0.11 & 0.968 $\pm$ 0.03 & 0.762 $\pm$ 0.11 & 0.6 & 5.0 & 10 \\ 
& & DP-SGD & 0.960 $\pm$ 0.01 & 0.909 $\pm$ 0.02 & 0.712 $\pm$ 0.12 & 0.960 $\pm$ 0.01 & 0.811 $\pm$ 0.08 & 0.8 & 5.0 & 5 \\ 
& & DP-SGD & 0.991 $\pm$ 0.01 & 0.846 $\pm$ 0.01 & 0.200 $\pm$ 0.11 & 0.991 $\pm$ 0.01 & 0.319 $\pm$ 0.11 & 1.5 & 5.0 & 1\\ [0.5ex] 
\cline{2-11}
& SAGE & SGD & 0.985 $\pm$ 0.01 & 0.946 $\pm$ 0.01 & 0.757 $\pm$ 0.04 & 0.985 $\pm$ 0.01 & 0.856 $\pm$ 0.02 & - & - & -   \\ 
& & DP-SGD & 0.972 $\pm$ 0.01 & 0.932 $\pm$ 0.02 & 0.767 $\pm$ 0.09 & 0.972 $\pm$ 0.01 & 0.854 $\pm$ 0.06 & 0.6 & 5.0 & 10 \\ 
& & DP-SGD & 0.973 $\pm$ 0.02 & 0.928 $\pm$ 0.02 & 0.738 $\pm$ 0.09 & 0.973 $\pm$ 0.02 & 0.835 $\pm$ 0.06 & 0.8 & 5.0 & 5 \\ 
& & DP-SGD & 0.951 $\pm$ 0.07 & 0.841 $\pm$ 0.02 & 0.402 $\pm$ 0.30 & 0.951 $\pm$ 0.07 & 0.493 $\pm$ 0.24 & 1.5 & 5.0  & 1 \\ [0.5ex]
\hline\hline  
\parbox[t]{5mm}{\multirow{4}{*}{\rotatebox[origin=c]{90}{\textit{Molbace}}}} & GCN & SGD & 0.743 $\pm$ 0.00 & 0.655 $\pm$ 0.02 & 0.511 $\pm$ 0.03 & 0.820 $\pm$ 0.01 & 0.629 $\pm$ 0.02 & - & - & - \\ 
& & DP-SGD & 0.699 $\pm$ 0.01 & 0.670 $\pm$ 0.01 & 0.723 $\pm$ 0.02 & 0.608 $\pm$ 0.01 & 0.660 $\pm$ 0.01 & 0.5 & 5.0 & 20 \\ 
& & DP-SGD & 0.688 $\pm$ 0.01 & 0.609 $\pm$ 0.01 & 0.412 $\pm$ 0.01 & 0.834 $\pm$ 0.01 & 0.552 $\pm$ 0.01 & 0.6 & 5.0 & 10 \\
\cline{2-11}
  & GAT & SGD & 0.781 $\pm$ 0.01 & 0.726 $\pm$ 0.02 & 0.691 $\pm$ 0.07 & 0.766 $\pm$ 0.06 & 0.721 $\pm$ 0.02  & - & - & - \\ 
& & DP-SGD & 0.747 $\pm$ 0.02 & 0.580 $\pm$ 0.02 & 0.333 $\pm$ 0.07 & 0.862 $\pm$ 0.03 & 0.475 $\pm$ 0.07 & 0.5 & 5.0 & 20 \\ 
& & DP-SGD & 0.692 $\pm$ 0.03 & 0.518 $\pm$ 0.04 & 0.153 $\pm$ 0.10 & 0.935 $\pm$ 0.04 & 0.248 $\pm$ 0.14 & 0.6 & 5.0 & 10 \\
\cline{2-11}
  & SAGE & SGD & 0.785 $\pm$ 0.00 & 0.654 $\pm$ 0.01 & 0.484 $\pm$ 0.02 & 0.848 $\pm$ 0.01 &  0.616 $\pm$ 0.01 & - & - & - \\ 
  & & DP-SGD & 0.717 $\pm$ 0.00 & 0.620 $\pm$ 0.01 & 0.901 $\pm$ 0.00 & 0.299 $\pm$ 0.01 & 0.448 $\pm$ 0.02 & 0.5 & 5.0 & 20 \\ 
  & & DP-SGD & 0.701 $\pm$ 0.00 & 0.550 $\pm$ 0.01 & 0.262 $\pm$ 0.00 & 0.879 $\pm$ 0.01 & 0.403 $\pm$ 0.01 & 0.6 & 5.0 & 10 \\ 
\hline\hline
\end{tabular}
\caption{Summary of our experimental evaluation on four datasets: \textit{Synthetic}, \textit{Fingerprints}, \textit{ECG}, and \textit{Molbace} with different network types. We report results with SGD and DP-SGD training as well as varying privacy budgets $\varepsilon$. The scores are evaluated on the test sets with a standard deviation based on five independent runs. We find that our models achieve high performance when using our proposed DP-SGD training method. The performance decreases slowly when increasing privacy guarantees.}
\label{tab:results}
\end{table*}


\begin{figure}
    \centering
    \includegraphics[width=0.6\textwidth]{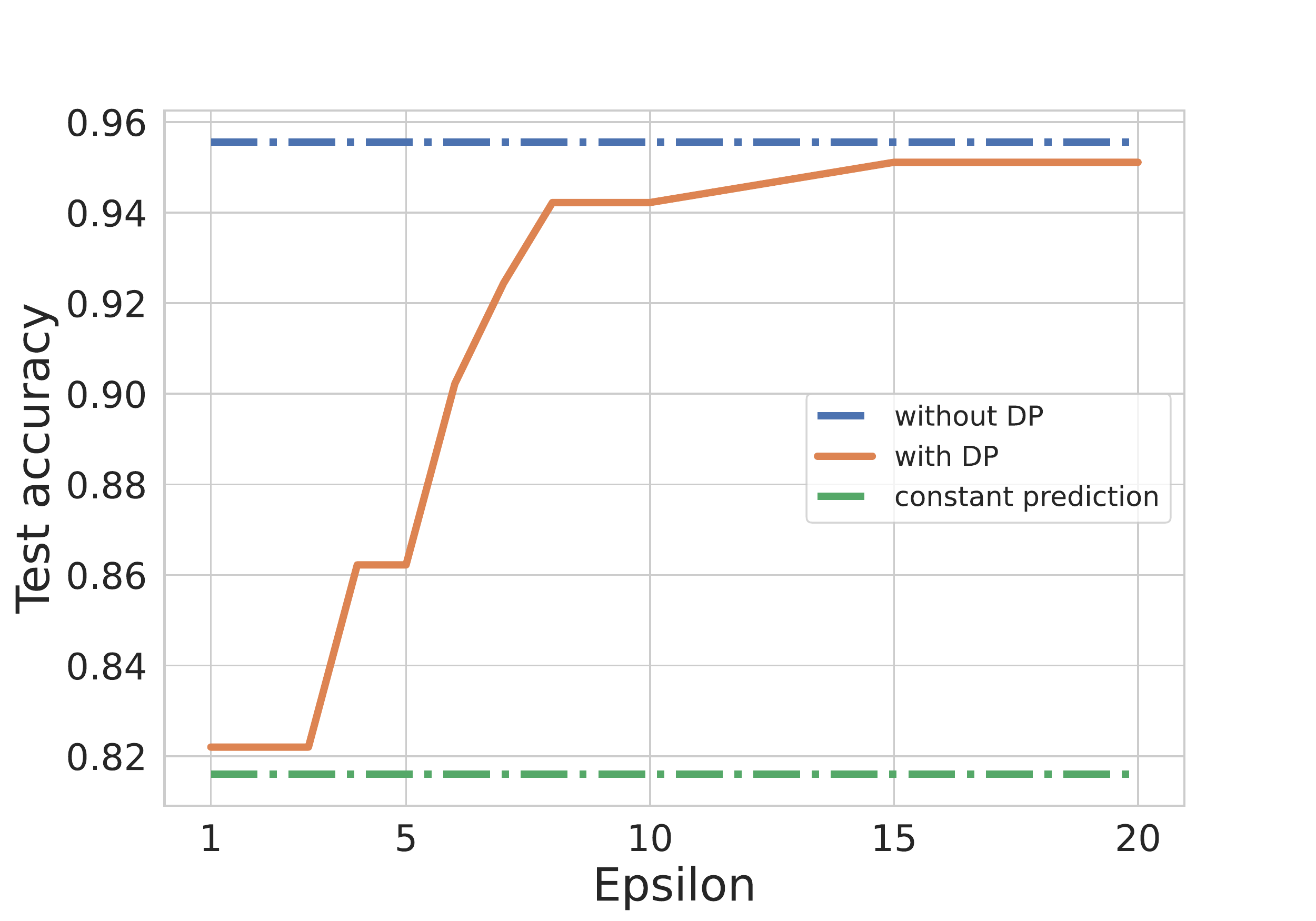}
    \caption{Impact of epsilon on test accuracy on ECG dataset. The performance increases with larger $\varepsilon$ values and looser privacy guarantees. We evaluated $\varepsilon$ values in $\{1,2,\ldots,10,15,20\}$}
    \label{fig:epsilon_ecg}
\end{figure}

In this section, we evaluate our results, compare DP-SGD training with standard SGD training and show the impact of different privacy budgets on model performances. The results achieved on the four datasets are summarised in Table \ref{tab:results}. 

\paragraph{Summary of Results} For all datasets, we observe similar behaviour, namely a correlation between stronger privacy budgets and diminished model performance. Although this phenomenon is -- in general -- an unavoidable, information-theoretic consequence of the trade-off between privacy and utility, the individual models exhibit different behaviour with regards to their individual tolerances towards the amount of Gaussian noise added for DP-SGD, as well as the tolerances towards gradient clipping. For instance, for the \textit{synthetic} dataset, an $\varepsilon$ value of $5$ does not lead to accuracy loss, whereas for the \textit{Molbace} dataset, a privacy budget of $\varepsilon=10$ already results in diminished model accuracy. 
Interestingly, the performance of DP-SGD training is overall not substantially influenced by the choice of GNN architecture (GCN, GAT or GraphSAGE). We observe high performance and similar convergence rates for all architectures, indicating the robust performance of DP-SGD training. For a comparison of the training behaviours please see our Figure in the supplementary material. 

For all models, we observe an increased inter-run variability with stronger privacy guarantees. This behaviour is reflected in the higher standard deviations reported in Table \ref{tab:results}, and we attribute this phenomenon to the increased randomness injected by the DP mechanism.

Exemplarily, we visualise the impact of a stronger privacy guarantee on the performance on the \textit{ECG} dataset in Figure \ref{fig:epsilon_ecg}. Given that the dataset is highly imbalanced, a constant prediction (marked by the lower dashed green line in Figure \ref{fig:epsilon_ecg}) would result in an approximate test accuracy of $81.6\%$. We examine the dependency of the results on the choice of $\varepsilon$ and report the different performances. With a very strong privacy guarantee (corresponding to a low $\varepsilon$ value), the performance of the network is barely better than a constant prediction. The looser the privacy guarantee (larger $\varepsilon$ value) the better the performance; for a very loose $\varepsilon$ the results reach non-DP performance. Interestingly, for some models we observe identical performance between DP-SGD and normal training, e.g. \textit{Fingerprint-GCN}, where the DP-SGD model (privacy budget of $\varepsilon=5$) reaches slightly higher performance then the normal training, see Table \ref{tab:results}; this beneficial effect can be attributed to the regularising effects of gradient norm bounding and noise injection, indicating that -- within certain constraints -- DP training can go hand-in-hand with excellent overall model performance and generalisability. 

\begin{figure}
    \centering
    \includegraphics[width=0.6\textwidth]{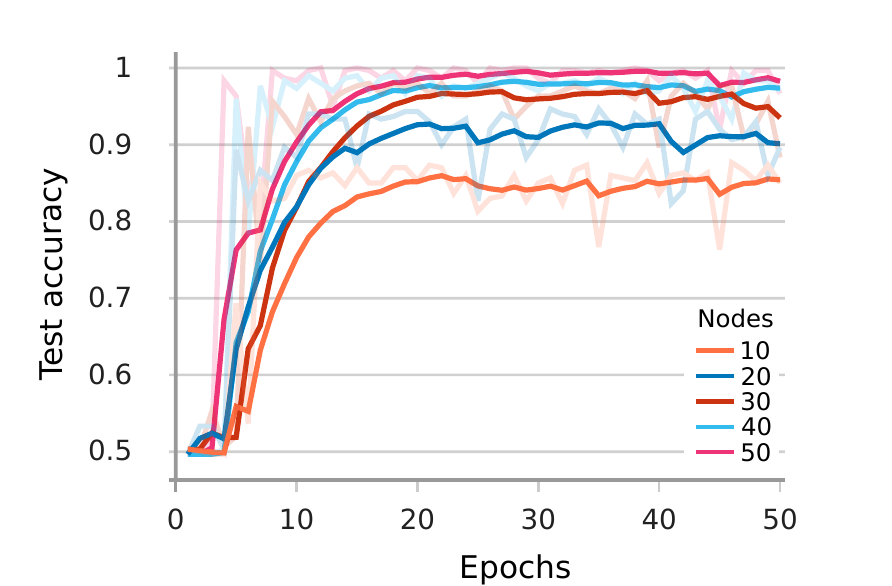}
    \caption{Impact of graph size to performance under DP: Increasing graph sizes result in better performance and faster convergence. The privacy guarantees are set to $\varepsilon=2.3$}
    \label{fig:node_size_impact}
\end{figure}

\paragraph{Scalability}
In order to investigate the scalability of our approach, we vary the size of the created Erd\H os-Rényi graphs in the \textit{synthetic} dataset between 10 and 500 nodes per graph. Figure \ref{fig:node_size_impact} shows the impact of the graph size on the performance under DP using a three-layer GCN and $\varepsilon=2.3$ . We visualised the performances of graph sizes between 10 and 50 nodes and find that performance improves with increasing graph size in these ranges. Beyond 50 nodes, the performance remains constantly high, which is why these plots were not included in Figure \ref{fig:node_size_impact}. This behaviour indicates a strong performance of our model across varying graph sizes; i.e. robust scalability. \\

\paragraph{Explainability}

\begin{figure}
    \centering
    \includegraphics[width=0.49\textwidth]{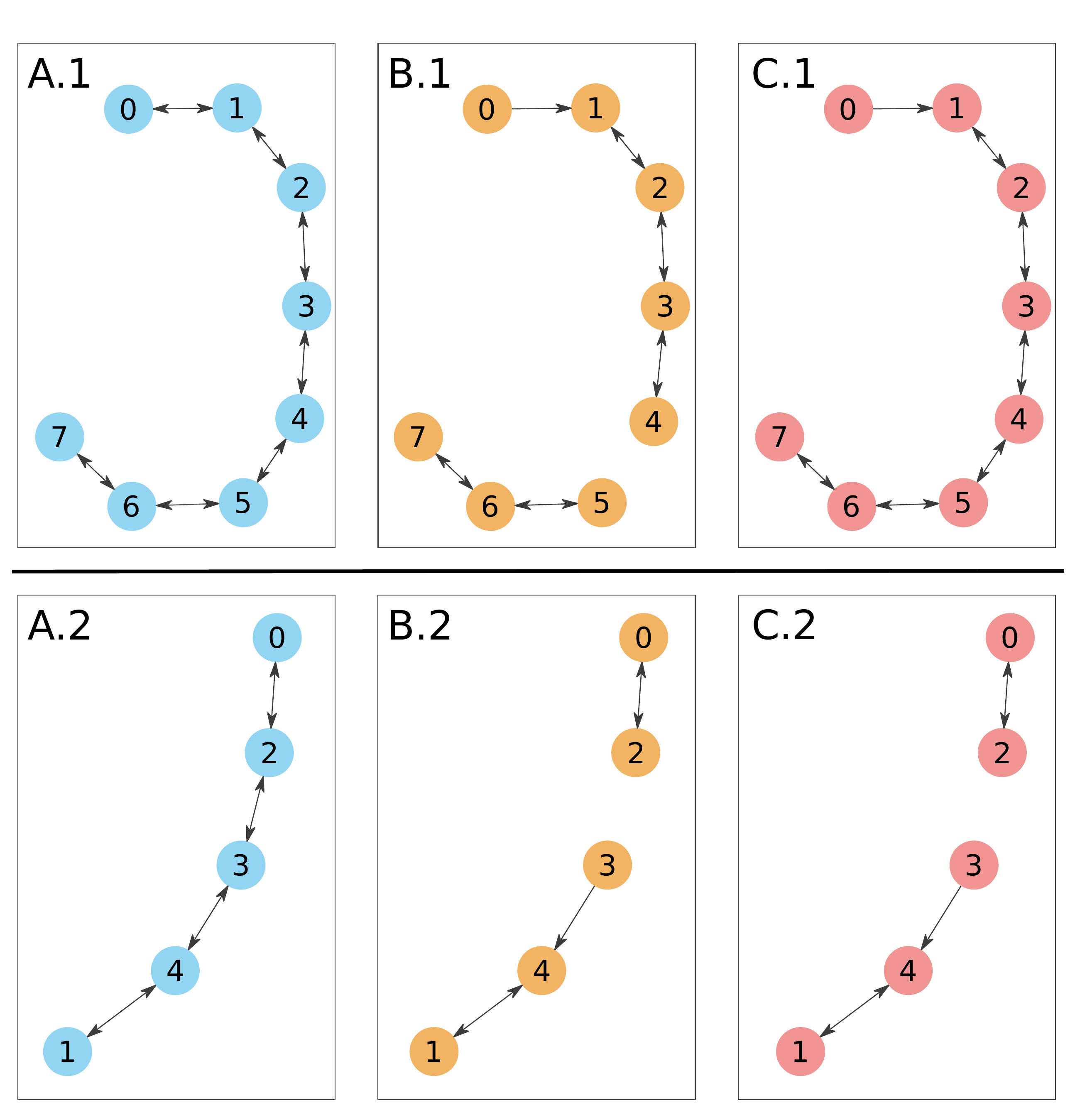}
    \caption{Visualisation of two GNNExplainer examples. The original graph (\textbf{A}) is shown in blue, the resulting graph from the GNNExplainer and the model trained with SGD in orange (\textbf{B}) and with DP-SGD in red (\textbf{C}). In the example in the upper row (\textbf{1}) the two graphs (\textbf{B}) and (\textbf{C}) differ slightly, whereas in the lower example (\textbf{2}) both GNNExplainer graphs (\textbf{B}) and (\textbf{C}) are equal, meaning that the two models consider the same edges to be relevant. The privacy budget for the models trained with DP-SGD was set to $\varepsilon$=5.}
    \label{fig:gnn_expl}
\end{figure}

The interpretability of GNNs is a challenging and frequently discussed task in research. Recently, approaches like the GNNExplainer \cite{ying2019gnn} formalised methods which can be used to interpret the results of trained GNNs. We make use of this method to interpret the differences in learned representations between models trained with DP-SGD and non-private SGD and visualise the results in Figure \ref{fig:gnn_expl}. The GNNExplainer is an approach for post-hoc interpretation of predictions generated by a trained GNN. It is used to identify which edges in the graph represent essential connections for a prediction of the network, thus indicating nodes important for the final prediction. \mbox{GNNExplainer} prunes the original graph to only contain the nodes and edges with the highest impact on the model prediction. We apply the GNNExplainer to our results on the \textit{Fingerprints} dataset, comparing a GCN model trained with standard SGD and three GCN models trained with DP-SGD with $\varepsilon=5$, $\varepsilon=1$ and $\varepsilon=0.5$. We set the GNNExplainer threshold for edge importance to $0.2$. Qualitatively, we observe that the GNNExplainer results of the DP models and the standard models appear very similar, if not identical for some examples, see Figures \ref{fig:gnn_expl} and supplementary material. In these Figures, (\textbf{A}) visualises an example of an original graph from the \textit{Fingerprints} dataset, containing all edges. Figures (\textbf{B}) and (\textbf{C}) show the pruned graphs for SGD and DP-SGD training, respectively. In the lower example (\textbf{2}) in Figure \ref{fig:gnn_expl}, both GNNExplainer graphs are identical (almost identical in the upper row), showing that in both models the same edges and nodes have a high impact on the models' predictions. This indicates that the feature importance is the same (or almost the same) between both models and that the feature importance is not compromised by the privacy guarantees achieved through DP training. 

\begin{table}
\footnotesize
\begin{center}
\begin{tabular}{lccc} 
$\varepsilon$ & ROC AUC & IoU(orig., DP) & \textbf{IoU(DP, non-DP)} \\ [0.5ex] 
 \midrule
 5.0 & 0.863 & 0.652 & \textbf{0.765} \\ 
 1.0 & 0.844 & 0.592 & \textbf{0.736} \\
 0.5 & 0.796 & 0.617 & \textbf{0.609} \\
[1ex] 
\end{tabular}
 \caption{Mean IoU scores of ten test samples from the \textit{Fingerprint} dataset for comparing edges between the original graph, the GNNExplainer graph of the model trained with SGD, and the GNNExplainer graph of the model trained with DP-SGD. The IoU between the original graph and the non-DP graph is \textbf{$0.739$}. The intersection between the DP and the non-DP graphs decreases with a smaller $\varepsilon$ value which corresponds to smaller ROC AUC results.} 
\label{tab:GNNExplainer}
\end{center}
\end{table}

To provide a quantitative estimation of GNNExplainer similarity of our results, we propose and use an Intersection over Union (IoU) score, measuring the pair-wise overlap of edges in the three resulting graphs. 
The IoU score of two graphs $A$ and $B$ is defined as follows:
\begin{equation}
    \frac{|\mathcal{E}_A \cap \mathcal{E}_B|}{|\mathcal{E}_A \cup \mathcal{E}_B|},
\end{equation}
where $\mathcal{E}_X$ represents the set of all edges in Graph $X$ and $\mid \cdot \mid$ denotes the cardinality of a set. 
Table \ref{tab:GNNExplainer} summarises the results of the mean IoU values between the original graph and the GNNExplainer graph based on training with DP, and the two resulting GNNExplainer graphs from DP-SGD and SGD training. The IoU score of the original graph and the GNNExplainer graph of the model trained with standard SGD is $0.739$ for all graphs. We compare the overlap between the graphs with the model performance, reported by the ROC AUC score. We find a high IoU score for DP vs. non-DP models, which is in line with the GNNExplainer plots we observe in Figure \ref{fig:gnn_expl}. Moreover, we observe that our GNNExplainer IoU score of the DP and the non-DP models slightly decreases with a smaller $\varepsilon$ and smaller ROC AUC scores, see Table \ref{tab:GNNExplainer}. The increase in the IoU score between the original model and the DP model with $\varepsilon=0.5$ most likely only indicates that the DP trained model with $\varepsilon=0.5$ considers more edges as relevant than the model trained with $\varepsilon=1.0$. These qualitative and quantitative GNNExplainer results indicate that our proposed DP graph classification models exhibit strong and similar inductive biases compared to \say{normal} GNNs while preserving privacy guarantees.

%% file: 06_conclusion.tex
\section{Discussion}
\label{sec:conclusion}
\paragraph{Conclusion} Our work introduces and evaluates differentially private graph classification, a formal method to offer quantifiable privacy guarantees in applications where sensitive data of individuals is represented as a whole graph. Such contexts include medical data (as shown in our ECG classification example), where DP can enable training of machine learning models while maintaining both regulatory compliance and adherence to ethical standards mandating the protection of health-related data. Our experiments on benchmark and real-world datasets demonstrate that the training of GNNs for graph classification is viable with high utility and tight privacy guarantees. Expectedly, we observe a privacy-performance trade-off, whereby a decrease in the value of $\varepsilon$ results in a decline in the accuracy of the model, as demonstrated in Figure \ref{fig:epsilon_ecg}.

Additionally, we investigate the utilisation of explainability techniques to compare the representations learned by models trained with SGD and DP-SGD. The application of the GNNExplainer indicates that models trained with DP-SGD learn similar relevant representations to the non-privately trained models. To quantitatively demonstrate the results of the GNNExplainer, we calculated an IoU score on the edges considered important by the technique between the resulting graphs. We observe an overall high IoU with a slight decline in overlap with tighter privacy guarantees, indicating that -- as expected -- the high levels of noise required to achieve such guarantees eventually become detrimental to learning.

\paragraph{Limitations and future work} 
Inherent to the concept of differential privacy in machine learning is a performance-to-privacy trade-off. While our experiments visually illustrate the implications of the trade-off and provide insight into its practical importance in the context of machine learning on graphs, the actual relationship between privacy and accuracy is highly task- and user-specific \cite{dwork2019differential,cummings2021need}. Therefore, we note that one can interpret the value of $\varepsilon$ as an additional design-parameter that needs to be optimised for in order to minimise the adverse effects that DP can have on performance in the context of graph classification (or most other learning tasks in general).

While the GNNExplainer concept can provide initial clues to interpret and explain GNN training and the intrinsic differences between models trained with SGD and DP-SGD, it is only an initial step towards full explainability and interpretability. We consider this to be a highly relevant and an interesting direction for future research. In particular, we aim to investigate the effects of differentially private GNN learning on adversarial robustness of the model. We hypothesise that -- similarly to Euclidean settings -- \cite{naseri2020toward,lecuyer2019certified} DP should have a mitigating effect against attacks that diminish the utility of the trained model in the context of machine learning on graphs.

In our experiments we utilize a limited set of standard model architectures (GCN, GraphSAGE, GAT). Evidently, more sophisticated architectures have been designed and deployed to real world problems. As our proposed approach is general, we assume that an extension to such advanced graph learning models is natural and should exhibit similar behaviour.

\paragraph{Discussion of potential societal impacts} 
We do not foresee any specific negative social impact of our work. To the contrary, we strongly believe that the implementation of formal techniques for privacy preservation like DP in the setting of GNN training will mitigate the risks of using sensitive data in ML tasks. In the case of medical data (as in the \textit{ECG} dataset example), we believe the utilisation of privacy preserving methods to also hold positive effects in terms of encouraging data owners (such as patients) to make their data accessible for research purposes. Evidently, such implementations must go hand in hand with educating potential stakeholders in the correct application of DP mechanisms, including the appropriate choice of parameters like $\varepsilon$. 
In this work, we rely exclusively on public datasets collected with informed consent or with approval of institutional review boards wherever applicable; thus, the risk of potential leakage of sensitive information during our experiments has been appropriately considered.

%% file: 07_appendix.tex
\section*{Supplementary Material}

\section{Model Architectures and Training \\Parameters}
We here provide a brief overview of the model architectures we used for our experiments as well all corresponding hyperparameters. For each dataset, we utilised a different model architecture which we determined through hyperparameter searches. Table \ref{tab:sub:architectures} summarises the hyperparameters used for each experiment. For SGD training on the \textit{Molbace} dataset we used a cyclic learning rate scheduler with a defined lower and upper learning rate as described in Table \ref{tab:sub:architectures}. We did not observe increased performance when applying the cyclic learning rate scheduler to DP-SGD training on this dataset, which is why we did not utilise a learning rate scheduler for those applications. All models are implemented using \textit{PyTorch} \cite{NEURIPS2019_9015} and \textit{PyTorch Geometric} \cite{Fey/Lenssen/2019}. For all binary models we used the Binary Cross Entropy Loss, for the non-binary classification task of the \textit{Fingerprint} dataset we use the Cross Entropy Loss.

\paragraph{Synthetic Dataset} For the \textit{synthetic} dataset we use a Neural Network (NN) with four layers of the corresponding graph convolutions (GCN, GAT, or GraphSAGE), followed by a Mean Pooling layer and two linear layers. Each convolutional layer, as well as the first linear layer, is followed by a ReLU activation function. The hidden channels of the graph convolutional layers are $200, 400, 800, 1600$ and the fully connected layers at the end of the network have $265$ hidden channels. For non-DP training, we used a learning rate of $0.05$ for the GCN network, $0.1$ for GAT and $0.08$ for GraphSAGE and a batch size of $24$. For all DP-SGD training runs we use an $L_2$ clipping bound of $3.0$ and noise multipliers in $\{1.0, 2.0, 2.2 \}$.

\paragraph{Fingerprint Dataset} For the experiments on the \textit{Fingerprint} dataset we use a NN architecture with an Instance Normalisation layer followed by three layers of the corresponding graph convolution (GCN, GAT, or GraphSAGE), a Max Pooling layer and two linear layers. Each convolutional layer, as well as the first linear layer, is followed by a ReLU activation function. The hidden channels of the graph convolutional layers are $256, 512, 1024$ and the fully connected layers at the end of the network have $265$ hidden channels. For the non-DP training we used a learning rate of $0.08$ for GCN, GAT and GraphSAGE networks and a batch size of $64$. DP-SGD training was performed with different noise multipliers in $\{1.0, 1.8, 2.3\}$ and the same $L_2$ Clip of $3.0$. The learning rates differ with the type of graph convolution and can be found in Table \ref{tab:sub:architectures}.

\paragraph{ECG Dataset} The experiments on the \textit{ECG} Dataset were performed with a model architecture consisting of two graph convolutional layers, followed by a Max Pooling layer and three linear layers. Each convolutional layer and each linear layer is followed by a ReLU activation function. The hidden channels of the graph convolutional layers are $256, 512$ and of the linear layers $128, 56, 24$. In the non-DP training all convolutional layers and the Max Pooling layer are followed by a Dropout Layer with dropout probability $0.2$. We removed all Dropout Layers for the DP-SGD training, because the added noise intrinsic to the algorithm already functions as regularisation. Learning rate and batch size of the SGD training were set to $0.05$ and $24$, correspondingly for all models. The learning rates for DP-SGD training runs depends on the graph convolution and can be found in Table \ref{tab:sub:architectures}.

\paragraph{Molbace Dataset} For the experiments on the \textit{Molbace} dataset we utilise a NN with an Instance Normalisation layer, followed by three graph convolution layers, each followed by another Instance Normalisation layer and a ReLU activation function. The convolutional layers are followed by a Mean Pooling layer and two linear layers with $512$ hidden channels. For the non-DP training we use a Batch Size of 64 and a cyclic learning rate scheduler with upper and lower learning rates noted in Table \ref{tab:sub:architectures}. The learning rates were determined using a learning rate finder. For the DP-SGD training we use a Batch Size of $24$ and a learning rate of $0.1$ for all models. We did not use a learning rate scheduler for DP-SGD training, since it did not improve model performance. 

\begin{table*}[h]
    \centering
    \begin{tabular}{lllccc}
         Dataset & Model & Optimiser & Learning Rate & Batch Size & Scheduler \\
         \hline
         \textit{Synthetic} & GCN & SGD & 0.05 & 24 & - \\
                            &  & DP-SGD & 0.1 & 24 & - \\
        \cline{2-6}
        & GAT & SGD & 0.1 & 24 & - \\
        &     & DP-SGD & 0.4 & 24 & - \\
        \cline{2-6}
        & SAGE & SGD & 0.04 & 24 & - \\
        &     & DP-SGD & 0.2 & 24 & - \\
        \hline
         \textit{Fingerprint} & GCN & SGD & 0.08 & 64 & - \\
         & & DP-SGD & 0.2 & 64 & - \\
         \cline{2-6}
         & GAT & SGD & 0.08 & 64 & - \\
         & & DP-SGD & 0.2 & 64 & - \\
         \cline{2-6}
         & SAGE & SGD & 0.08 & 64 & - \\
         & & DP-SGD & 0.1 & 64 & - \\
         \hline
         \textit{ECG} & GCN & SGD & 0.05 & 24 & - \\
         & & DP-SGD & 0.12 & 24 & - \\
         \cline{2-6}
         & GAT & SGD & 0.05 & 24 & - \\
         & & DP-SGD & 0.15 & 24 & - \\
         \cline{2-6}
         & SAGE & SGD & 0.05 & 24 & - \\
         & & DP-SGD & 0.1 & 24 & - \\
         \hline
         \textit{Molbace} & GCN & SGD & $7e^{-3}$ to $7e^{-2}$ & 64 & cyclic \\
         & & DP-SGD & 0.1 & 24 & - \\
         \cline{2-6}
         & GAT & SGD & $5e^{-3}$ to $5e^{-2}$ & 64 & cyclic \\
         & & DP-SGD & 0.1 & 24 & - \\
         \cline{2-6}
         & SAGE & SGD & $7e^{-3}$ to $7e^{-2}$ & 64 & cyclic \\
         & & DP-SGD & 0.1 & 24 & - \\
         \hline
    \end{tabular}
    \caption{Overview of hyperparameters for all datasets and experiments. We performed manual and grid search hyperparameter tuning.}
    \label{tab:sub:architectures}
\end{table*}

\section{Performances of Different Model \\Architectures}
\begin{figure}
    \centering
    \includegraphics[width=0.5\textwidth]{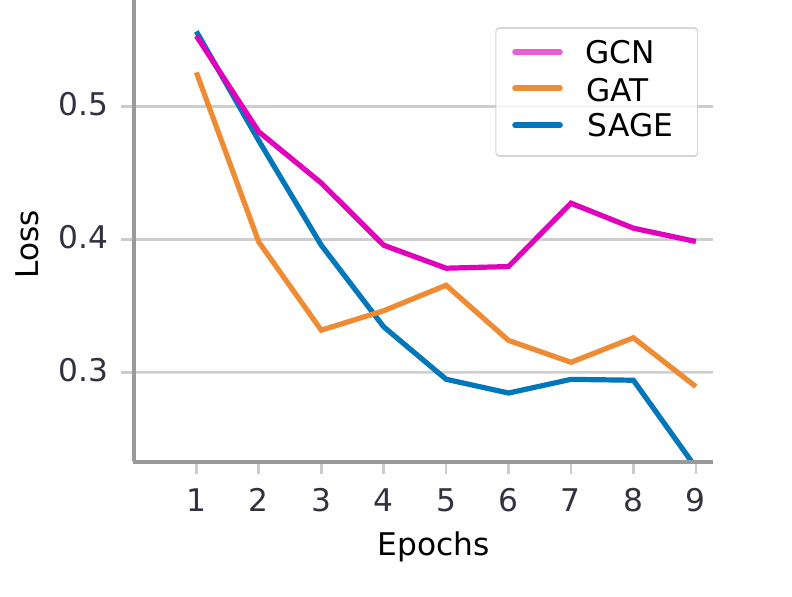}
    \caption{Training loss curves of the three GNN models on the \textit{ECG} dataset using DP-SGD with a noise multiplier of $0.6$ and an $L_2$ Clip of $5.0$, where $\varepsilon=10$.}
    \label{fig:sub:train_losses_ecg}
\end{figure}

\begin{figure}
    \centering
    \includegraphics[width=0.5\textwidth]{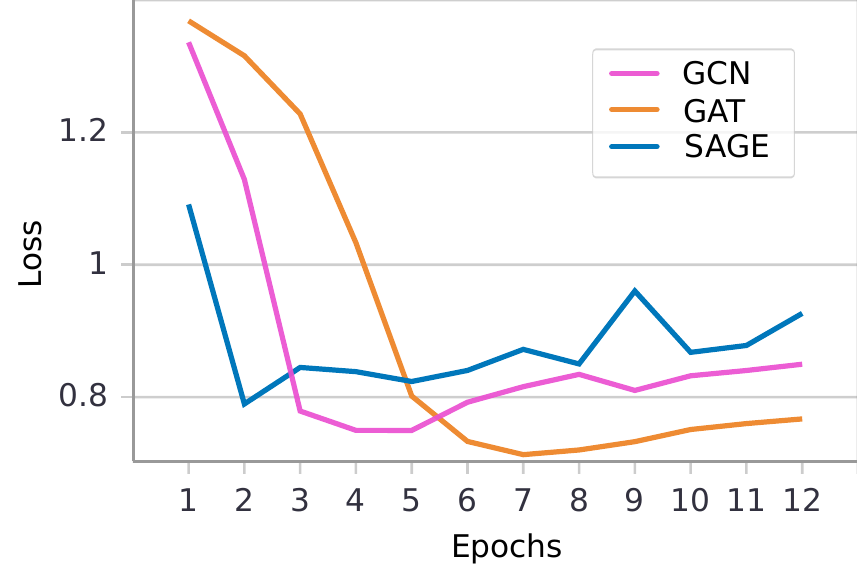}
    \caption{Training loss curves of the three GNN models on the \textit{Fingerprint} dataset using DP-SGD with a noise multiplier of $1.0$ and an $L_2$ Clip of $3.0$, where $\varepsilon=5$.}
    \label{fig:sub:train_losses_fingerprints}
\end{figure}

The comparable performance of the different GNN architectures we use for our experiments (GCN, GAT, and GraphSAGE) leads to the conclusion that DP-SGD training is independent from the type of graph convolution. For all datasets, we report similar performance of all models and we show two examples of respective training loss curves in Figure \ref{fig:sub:train_losses_ecg} and \ref{fig:sub:train_losses_fingerprints}. Figure \ref{fig:sub:train_losses_ecg} shows the training loss curves on the \textit{ECG} dataset for the DP-SGD training with noise multiplier $0.6$ and $L_2$ Clip $5.0$.


\section{Explainability using GNNExplainer}
\begin{figure}
    \centering
    \includegraphics[width=0.49\textwidth]{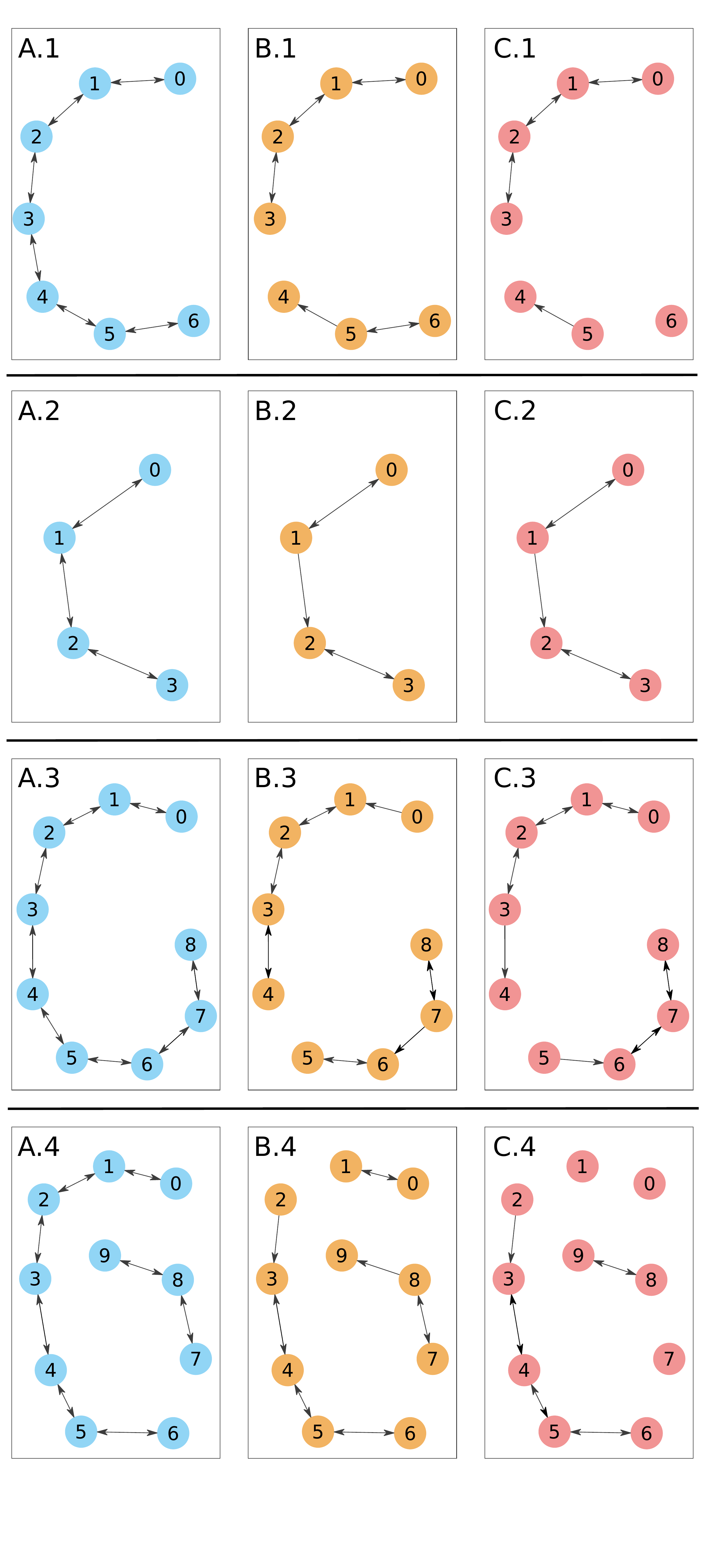}
    \caption{$\varepsilon=5$, fingerprints dataset, Visualisation of four GNNExplainer examples. The original graph (\textbf{A}) is shown in blue, the resulting graph from the GNNExplainer and the model trained with SGD in orange (\textbf{B}) and with DP-SGD in red (\textbf{C}). Both models were trained on the \textit{Fingerprint} dataset. The DP-SGD training was performed with a privacy budget of $\varepsilon=5$.}
    \label{fig:sub:gnnexplainer}
\end{figure}

As noted in section \ref{sec:results}, we applied the explainability technique GNNExplainer \cite{ying2019gnnexplainer} to our trained networks. Figure \ref{fig:sub:gnnexplainer} visualises four more examples of the original graph (\textbf{A}) in blue, the output of the GNNExplainer from the \say{normal} SGD training (\textbf{B}) in orange and the graph resulting from the GNNExplainer and the DP-SGD training (\textbf{C}) in red. All figures were created using our GCN model on the \textit{Fingerprint}  dataset. The DP model was trained with a privacy budget of $\varepsilon=5$. In the second example, \textbf{B.2} and \textbf{C.2} are equal, indicating that in both SGD and DP-SGD training the same graph edges are considered most relevant. In the other three examples in Figure \ref{fig:sub:gnnexplainer} there are minute differences in the GNNExplainer graphs, showing that slightly different edges have most impact on the model's prediction.

